\documentclass[runningheads]{llncs}
\usepackage{graphicx}

\usepackage{tikz}
\usepackage{comment}
\usepackage{amsmath,amssymb} 
\usepackage{color}
\usepackage{multirow}

\usepackage[ruled,vlined]{algorithm2e}

\begin{document}
\pagestyle{headings}
\mainmatter
\def\ECCVSubNumber{27}  


\title{WaveTransform: Crafting Adversarial Examples via Input Decomposition}

\titlerunning{WaveTransform ECCVW'20}
%

\author{Divyam Anshumaan\inst{1} \and
Akshay Agarwal\inst{1,2} \and Mayank Vatsa\inst{3} \and Richa Singh\inst{3}  \\
{ $^1$\{divyam17147,akshaya\}@iiitd.ac.in, $^3$\{mvatsa,richa\}@iitj.ac.in}}
\authorrunning{Anshumaan et al.}
%
\institute{$^1$IIIT-Delhi, India, $^2$Texas A\&M University, Kingsville, USA, $^3$IIT Jodhpur, India}

\maketitle

\begin{abstract}


Frequency spectrum has played a significant role in learning unique and discriminating features for object recognition. Both low and high frequency information present in images have been extracted and learnt by a host of representation learning techniques, including deep learning. Inspired by this observation, we introduce a novel class of adversarial attacks, namely `WaveTransform', that creates adversarial noise corresponding to low-frequency and high-frequency subbands, separately (or in combination). The frequency subbands are analyzed using wavelet decomposition; the subbands are corrupted and then used to construct an adversarial example. Experiments are performed using multiple databases and CNN models to establish the effectiveness of the proposed WaveTransform attack and analyze the importance of a particular frequency component. The robustness of the proposed attack is also evaluated through its transferability and resiliency against a recent adversarial defense algorithm. Experiments show that the proposed attack is effective against the defense algorithm and is also transferable across CNNs.

\end{abstract}

\section{Introduction}

\begin{figure}[t]
\begin{center}
   \includegraphics[width=0.925\linewidth]{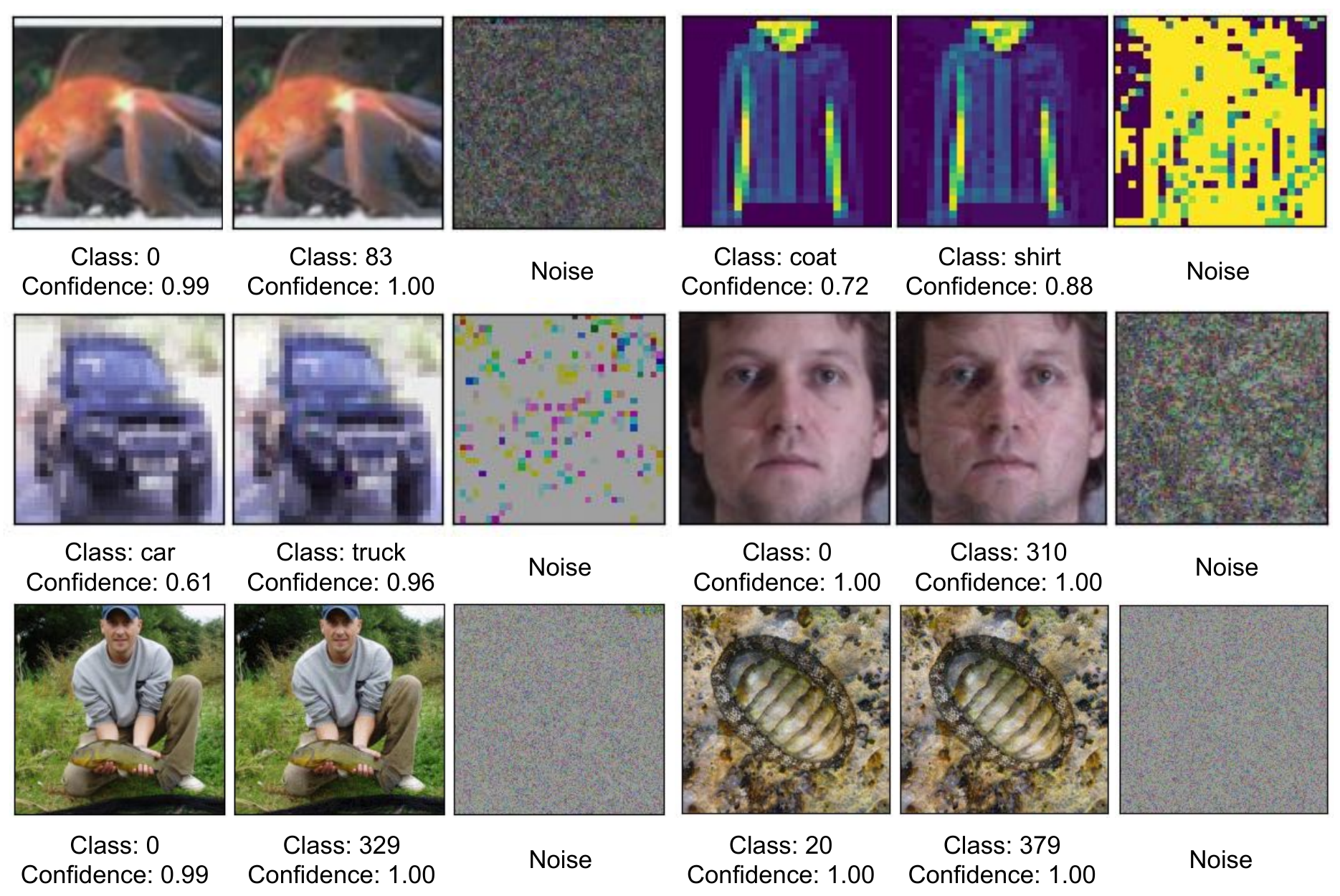}
\end{center}
\caption{Fooling of CNN model using the proposed attack on a broad range of databases including object recognition (Tiny ImageNet \cite{yao2015tiny}, ImageNet \cite{deng2009imagenet}, CIFAR-10 \cite{Krizhevsky2009LearningML}), face identification (Multi-PIE \cite{10.1016/j.imavis.2009.08.002}), and Fashion data classification (Fashion-MNIST \cite{xiao2017fashionmnist}). In each image set, the first image is the clean image, the second is an adversarial image, and the last is the adversarial noise. It can be clearly observed that the proposed attack is able to fool the networks with high confidence score.}
\label{fig:motive}
\end{figure}

Convolutional neural networks (CNNs) for image classification are known to utilize both high and low frequency information \cite{matthew2014visualizing}, \cite{wang2019high}. Goodfellow et al. \cite{goodfellow2014explaining} show that the CNN activations are sensitive towards high-frequency information present in an image. It is also shown that some neurons are sensitive towards the upper right stroke, while some are activated for the lower edge. Furthermore, Geirhos et al. \cite{geirhos2018imagenet} have shown that the CNN trained on ImageNet \cite{deng2009imagenet} are highly biased towards texture (high-frequency) and shape of the object (low-frequency). We hypothesize that if an attacker can manipulate the frequency information presented in an image, it can fool CNN architectures as well. With this motivation, we propose a novel method of adversarial example generation that utilizes the low-frequency and high-frequency information individually or in combination. To find the texture and shape information, a wavelet-based decomposition is an ideal choice which yields multi-resolution high-frequency and low-frequency images. Therefore, the proposed method incorporates wavelet decomposition to obtain multiple high and low-frequency images and adversarial noise is added to individual or combined wavelet components through gradient descent learning to generate an adversarial example. Since almost every CNN learns these kinds of features; therefore, the attack generated by perturbing the high frequency (edge) information makes it easily transferable to different networks. In brief, the key highlights of this research are: 



\begin{itemize}
    \item a novel class of adversarial example generation is proposed by decomposing the image into low-frequency and high-frequency information via wavelet transform;
    \item extensive experiments concerning multiple databases including ImageNet \cite{deng2009imagenet}, CIFAR-10 \cite{Krizhevsky2009LearningML}, and Tiny ImageNet \cite{yao2015tiny}, 
    \item multiple CNN models including ResNet \cite{he2015deep} and DenseNet \cite{huang2016densely} are used to showcase the effectiveness of the proposed WaveTransform;
    \item the robustness of the proposed attack is evaluated against a recent complex adversarial defense.
\end{itemize}

Fig. \ref{fig:motive} shows the effectiveness of the proposed attack on multiple databases covering color and gray-scale object images to face images. The proposed attack can fool the network trained on each data type with high confidence. For example, on the color object image (the first image of the top row), the model predicts the correct class (i.e., $0$) with confidence $0.99$, while, after the attack, the network misclassifies it to the wrong category (i.e., $83$) with confidence $1.00$.

\section{Related Work}

Adversarial generation algorithms presented in the literature can be divided into the following categories: (i) gradient-based, (ii) optimization-based, (iii) decision boundary-based, and (iv) universal perturbation. 

Goodfellow et al. \cite{goodfellow2014explaining} proposed a fast attack method that calculated the gradient of the image concerning the final output and pushed the image pixels in the direction opposite to the sign of the gradient. The adversarial noise vector can be defined as: $\eta = \epsilon sign(\triangledown_x J_\theta (x,l)) $, where $\epsilon$ controls the magnitude of perturbation, $\triangledown_x$ represents the gradient of image $x$ with respect to network parameters $\theta$. The perturbation vector $\eta$ is added in the image to generate the adversarial image. The above process is applied for a single step, which is less effective and can easily be defended \cite{kurakin2016adversarial+}. Therefore, several researchers have proposed the variant where the noise is added iteratively \cite{kurakin2016adversarial}, \cite{aleks2017deep}, and with momentum \cite{dong2018boosting}. Moosavi-Dezfooli et al. \cite{moosavi2016deepfool} have proposed a method that can transfer clean images from their decision boundaries to some other, belonging to a different class. The attacks are performed iteratively using a linear approximation of the non-linear decision boundary. Carlini and Wagner \cite{carlini2017towards} presented attacks by restricting the $L_{2}$ norm of an adversarial image. The other variant, such as $L_{\infty}$ and $L_{1}$, are also proposed; however, they are found to be less effective as compared to $L_{2}$. Similar to $L_{2}$ norm minimization, Chen et al. \cite{chen2018ead} have proposed the elastic norm optimization attack, which is the combination of $L_{2}$ and $L_{1}$ norm. Goswami et al. \cite{aaa2018goswami,ijcv2019goswami} presented several black-box attacks to fool the state-of-the-art face recognition algorithms. Later, both adversarial examples detection and mitigation algorithms are also presented in the paper. Agarwal et al. \cite{agarwal2020noise} shown the use of filtering operations in generating adversarial noise in a network-agnostic manner.

Other popular adversarial generation algorithms are based on generative networks \cite{xiao2018generating}, and EoT \cite{athalye2017synthesizing}. The application of an adversarial attack is not restricted to 2D object recognition but also explored for semantic segmentation \cite{xie2017adversarial}, 3D recognition \cite{xiang2019generating}, audio classification \cite{carlini2018audio}, text recognition \cite{gao2018black}, and reinforcement learning \cite{behzadan2017vulnerability}. Goel et al. \cite{goel2018smartbox} have developed an adversarial toolbox for the generation of adversarial perturbations and defense against them. The details of the existing algorithms can be found in the survey papers presented by Yuan et al. \cite{8611298} and Singh et al. \cite{singh2020robustness}.


\begin{figure}[t]
\begin{center}
   \includegraphics[width=1.0\linewidth]{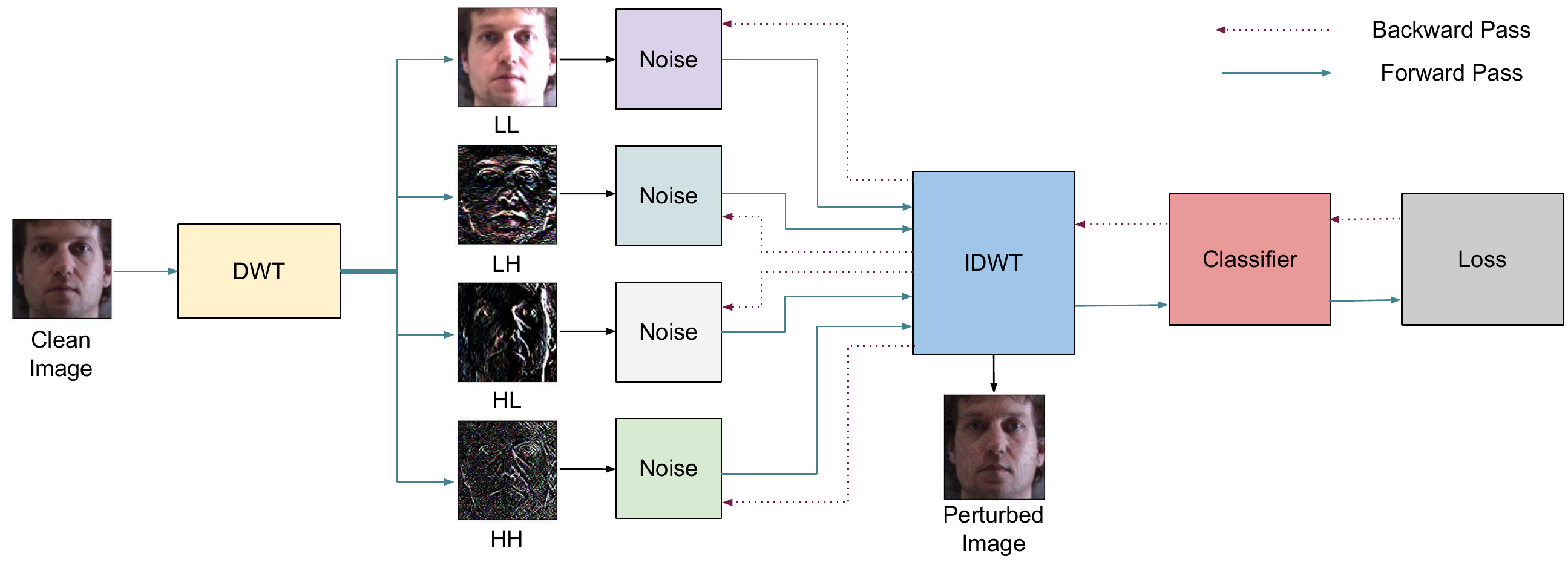}
\end{center}
\caption{Schematic diagram of the proposed \textit{`WaveTransform'} adversarial attack algorithm. DWT and IDWT are the forwards and inverse discrete wavelet decomposition. The noise is added to the desired wavelet subband and optimized to increase the loss of the network. LL represents the low pass subband. LH, HL, and HH represent the high-pass subbands in horizontal, vertical, and diagonal directions.}
\label{fig:algo}
\end{figure}

\section{Proposed WaveTransform Attack Algorithm}


Adversarial attacks generally modify the image in the spatial domain. In this research, we propose a new class of attack termed as WaveTransform where the image is first transformed into the frequency (scale) domain using wavelet decomposition. A digital image is composed of low frequency and high-frequency information, where the role of each frequency component might be different in its spatial representation. With this observation, high and low-frequency bands are perturbed such that the reconstructed image is an adversarial example but visually close to the clean example. The proposed attack can be defined using the following equation:
\begin{equation}\label{attack-constrain}
    min\;\alpha\|\mathcal{I}_{org} - \mathcal{I}_{pert}\|_{\infty} + \mathcal{L}(\mathcal{F}(\,\mathcal{I}_{pert}) \,, t) \,
\end{equation}
where, $\mathcal{I}_{org}, \mathcal{I}_{pert}\in[ \,0, 1] \,$ represent the clean and perturbed images, respectively. $\alpha$ is the loss term trade-off parameter, $\mathcal{L}$ is the classification loss function of the target CNN classifier $\mathcal{F}$, and $t$ is the target label. The aim is to find an adversarial image $\mathcal{I}_{pert}$ that maximizes the classification error for a target label while keeping the noise imperceptible to the human observer.

\begin{algorithm}[!t]
    \caption{Subband Updating (Proposed Adversarial Attack)}
    \textbf{Initialization}: 
    \begin{itemize}
        \item Let the selected subbands be expressed by $\theta$, for a particular image $\mathcal{I}_{j}$.
        \item Let the perturbed image be $\mathcal{I}_{j}^{'}\gets IDWT( \,\theta) \,$
        \item Let $l_{j}$ be the ground truth label of the image.
        \item Let $r$ be the number of random restarts taken, $k$ be the number of steps to optimize the objective function 
        \item Let $\gamma$ be the step size of the update and let $n$ be the minibatch size.
        \item Let the CNN model be expressed as $\mathcal{F}$.
        \item Let $\epsilon$ be the maximum amount of noise that may be added to $\mathcal{I}_{j}^{'}$, such that $\mathcal{I}_{j}^{'} \in [ \,\mathcal{I}_{j}-\epsilon, \mathcal{I}_{j}+\epsilon] \,$
    \end{itemize}
    \For{restarts in r}{
        Initialize $\mathcal{I}_{j}^{'}$ by adding random noise to $\mathcal{I}_{j}$ from range $[ \,-\epsilon, \epsilon] \,$ \\
        \For{steps in k}{
            Obtain subbands ($\theta$) by decomposing $\mathcal{I}_{j}^{'}$.
            \begin{center}
                $\theta_{low}, \theta_{high} \gets DWT( \,\mathcal{I}_{j}^{'}) \,$
            \end{center}
            Update subband(s) to maximize classification error by gradient ascent using the term:
            \begin{center}
                $\theta \gets \theta + \gamma( \,sign( \,\nabla_{\theta}\mathcal{L}( \,\mathcal{F}( \,IDWT( \,\theta_{low}, \theta_{high}) \,) \,, l_{j}) \,) \,$
            \end{center}
            \vskip 0.1cm
            $x\gets IDWT( \,\theta_{low}, \theta_{high}) \,$\\
            \vskip 0.1cm
            Project $x$ into valid range by clipping pixels and update $\mathcal{I}_{j}^{'}$ \;
            $\mathcal{I}_{j}^{'}\gets x$\;
            \vskip 0.1cm
            \If{$\mathcal{F}(x) \, \neq l_{j}$}{
                  Return  $\mathcal{I}_{j}^{'}$ 
            }
        }
    }
    Return $\mathcal{I}_{j}^{'}$
    \label{alg:two}
 \end{algorithm}

A discrete wavelet transform ($DWT$) is applied on $\mathcal{I}_{org}$ to obtain the $LL, LH$, $HL$, and $HH$ subbands, using low pass and high pass filters. The LL band contains low frequency information. Whereas LH, HL, and HH contain the high frequency information in horizontal, vertical, and diagonal directions, respectively. These subbands are then modified by taking a step in the direction of the sign of the gradient of the subbands concerning the final output vector. The image is then reconstructed with the modified subbands using an inverse discrete wavelet transform ($IDWT$), to obtain the desired image $\mathcal{I}_{pert}$.  As shown in Fig. \ref{fig:algo}, the attack is performed iteratively to find an adversarial image with minimal distortion. It is ensured that $\mathcal{I}_{pert}$ remains a valid image after updating its wavelet subbands by projecting the image back onto a $L_{\infty}$ ball of valid pixel values such that $\mathcal{I}_{pert}\in [ \,0,1] \,$. If the noise that can be added or removed, is already limited to $\epsilon$, we add another clipping operation limiting pixel values such that $\mathcal{I}_{pert}\in [ \,\mathcal{I}_{org}-\epsilon, \mathcal{I}_{org}+\epsilon] \,$. Since, in this setting, there is no need to minimize the added noise explicitly, we also fix the trade-off parameter to $\alpha=0$. Based on this, we propose our main method called Subband Updating, where particular subbands obtained by the discrete wavelet transform of the image are updated using projected gradient ascent. The proposed `WaveTransform' adversarial attack algorithm is described in Algorithm \ref{alg:two}.

\section{Experimental Setup}\label{section:experiments}
The experiments are performed using multiple databases and CNN models. This section describes the databases used to generate the adversarial examples, CNN models used to report the results and parameters for adversarial attack and defense algorithms.

\noindent \textbf{Databases:}
The proposed method is evaluated with databases comprising a wide range of target images: Fashion-MNIST (F-MNIST) \cite{xiao2017fashionmnist}, CIFAR-10 \cite{Krizhevsky2009LearningML}, frontal-image set of Multi-PIE \cite{10.1016/j.imavis.2009.08.002}, Tiny-ImageNet \cite{yao2015tiny}, and ImageNet \cite{deng2009imagenet}. Fashion-MNIST comprises low-resolution grayscale images of 10 different apparel categories. CIFAR-10 contains low-resolution RGB images of 10 different object categories. Multi-PIE database has high-resolution RGB images of 337 individuals and Tiny-ImageNet \cite{yao2015tiny} contains 10,000 images from over 200 classes from the ILSVRC challenge \cite{ILSVRC15}. To perform the experiments on ImageNet, the validation set comprising 50,000 images are used. These datasets also vary in color space, CIFAR-10 and Tiny-Imagenet contain color images while F-MNIST contains gray-scale images.

\begin{table}[!b]
    \begin{center}
    \caption{Architecture of the custom model used for Fashion-MNIST experiments. \cite{xiao2017fashionmnist}.}
    \label{tab:fmnist-model}
    \setkeys{Gin}{keepaspectratio}
    \resizebox*{0.5\textwidth}{\textheight} {
      \begin{tabular}{l|c|c} \hline
        \textbf{Layer Type}&\textbf{Output Size}&\textbf{Description}\\\hline
        Batch Norm 2D&28x28&channels 1, affine False\\\hline
        Conv 2D&28x28&5x5, 10, stride 1\\\hline
        Max Pool 2D&24x24&kernel 2x2\\\hline
        ReLU&23x23& - \\\hline
        Conv 2D&23x23&5x5, 20, stride 1\\\hline
        Max Pool2D&21x21&kernel 2x2\\\hline
        ReLU&20x20& - \\\hline
        Dropout 2D&20x20&Dropout prob 0.2\\\hline
        Flatten&400x1& Convert to a 1D vector\\\hline
        Linear&400x1&320, 10\\\hline
        Output&10x1& Final logits\\\hline
    \end{tabular}
    }
    \end{center}
 
\end{table}

\noindent \textbf{CNN Models and Implementation Details:} Recent CNN architectures with high classification performance are used for the experiments. For Multi-PIE, we use a ResNet-50 model \cite{he2015deep} pretrained on VGG-Face 2 \cite{Cao18} and an InceptionNet-V1 \cite{szegedy2014going} pretrained on CASIA-Webface \cite{yi2014learning}. For CIFAR-10, pretrained ResNet-50 \cite{he2015deep} and DenseNet-121 \cite{huang2016densely} are used, pretrained on the same. For Fashion-MNIST, a 10-layer custom CNN, as described in Table \ref{tab:fmnist-model}, has been used, and a pretrained ResNet-50 is used with Tiny-ImageNet and ImageNet. The standard models are fine-tuned, replacing the last layer of the network to match the number of classes in the target database and then iterating over the training split of the data for 30 epochs using the Adam \cite{kingma2014adam} optimizer with a learning rate of 0.0001 and batch size of 128. Standard train-validation-test splits are used for CIFAR-10, Fashion-MNIST, and Tiny-ImageNet databases. From the Multi-PIE database \cite{10.1016/j.imavis.2009.08.002}, 4753 training images, 1690 validation, and 3557 test images are randomly selected. All the models use images in the range $[0, 1]$, and the experimental results are summarized on the test split of the data, except for ImageNet and Tiny-ImageNet, where experimental results are reported on the validation split. 



\noindent \textbf{Attack Parameters:} \label{protocol}
In the experiments, each attack follows the same setting unless mentioned. Cross-entropy is used as the classification loss function $\mathcal{L}$. The SGD \cite{kiefer1952} optimizer is used to calculate the gradient of the subbands concerning the final logits vector used for classification. The experiments are performed using multiple different wavelet filters including Haar, Daubechies (db2 and db3), and Bi-orthogonal. Before computing the discrete wavelet transform, input data is extrapolated by zero-padding. Each attack runs for 20 iterations with 20 restarts, where the adversarial image is initialized with added random noise. This is referred to as random restarts by Madry et al. \cite{aleks2017deep}, where the attack algorithm finally returns the first valid adversarial image produced between all restarts. The maximum amount of noise that may be added to (or removed from) a clean image is fixed at $\epsilon=8.0/255.0$ in terms of $L_{\infty}$ norm for all the attacks. The step size of the subband update is fixed at $\gamma=0.05$.

\section{{Results and Observations}} 


This section describes the results corresponding to original and adversarial images generated via perturbing the individual or combined wavelet components. Extensive analysis has been performed to understand the effect of different filters with wavelet transformation. To demonstrate the effectiveness of the transformed domain attack, we have compared the performance with prevalent pixel-level attacks and recent steganography based attacks. We have also evaluated the transferability of the proposed attack and resiliency against a recent defense algorithm \cite{wang2019high}. 

\begin{figure*}[!t]
\begin{center}
   \includegraphics[width=1.0\linewidth]{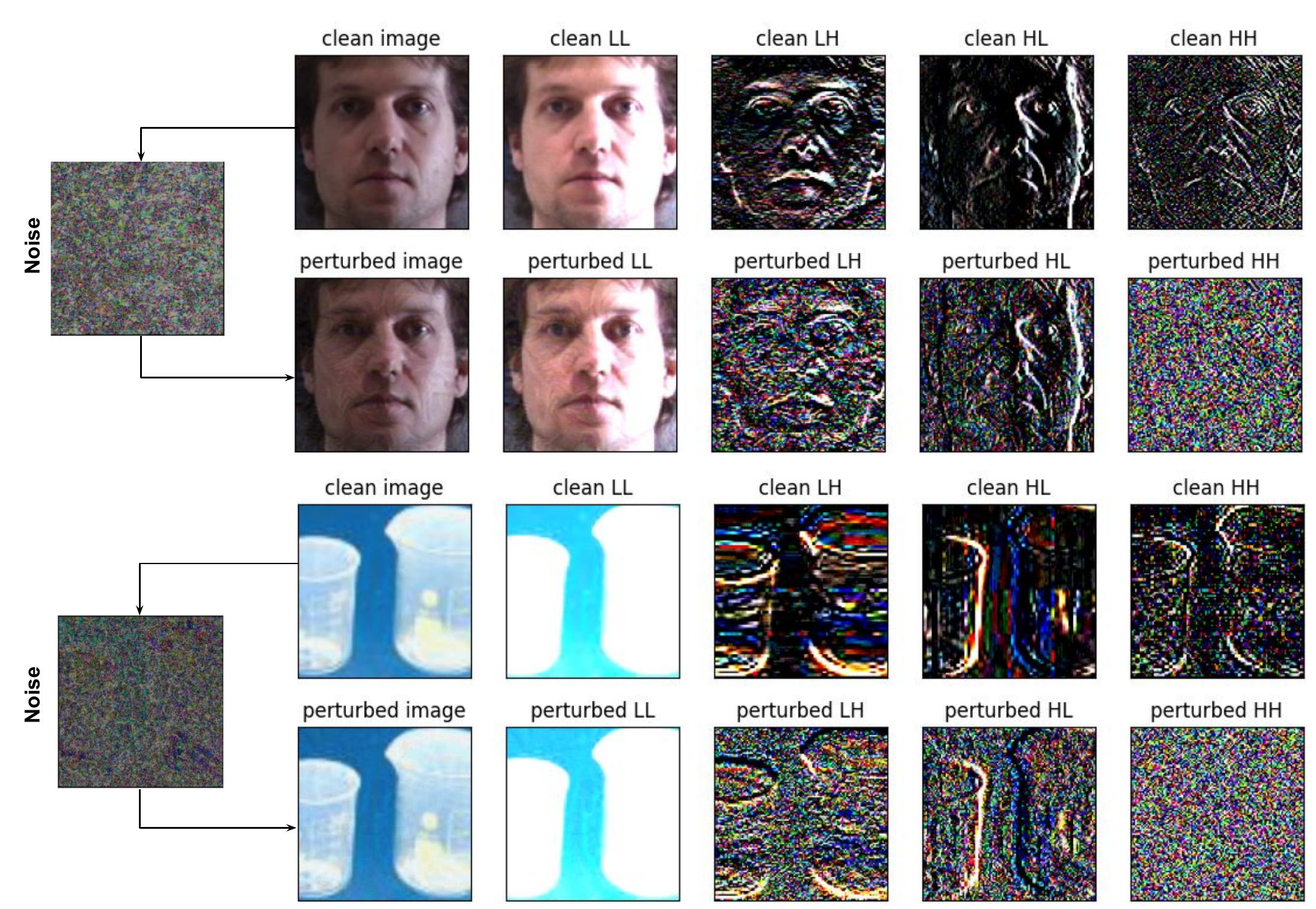}
\end{center}
\caption{Illustrating the individual wavelet components of the adversarial images generated using clean images from Multi-PIE \cite{10.1016/j.imavis.2009.08.002} and Tiny-ImageNet \cite{yao2015tiny} databases. While the adversarial images are visually close to the clean images; the individual high-frequency components ($LH$, $HL$, and $HH$) clearly show that the noise is injected to fool the system. 
The wavelet components corresponding to the $HH$ subband show the maximum effect of the adversarial noise.}
\label{fig:band-perturbed}
\end{figure*}

\begin{figure*}[!t]
\begin{center}
   \includegraphics[width=1.0\linewidth]{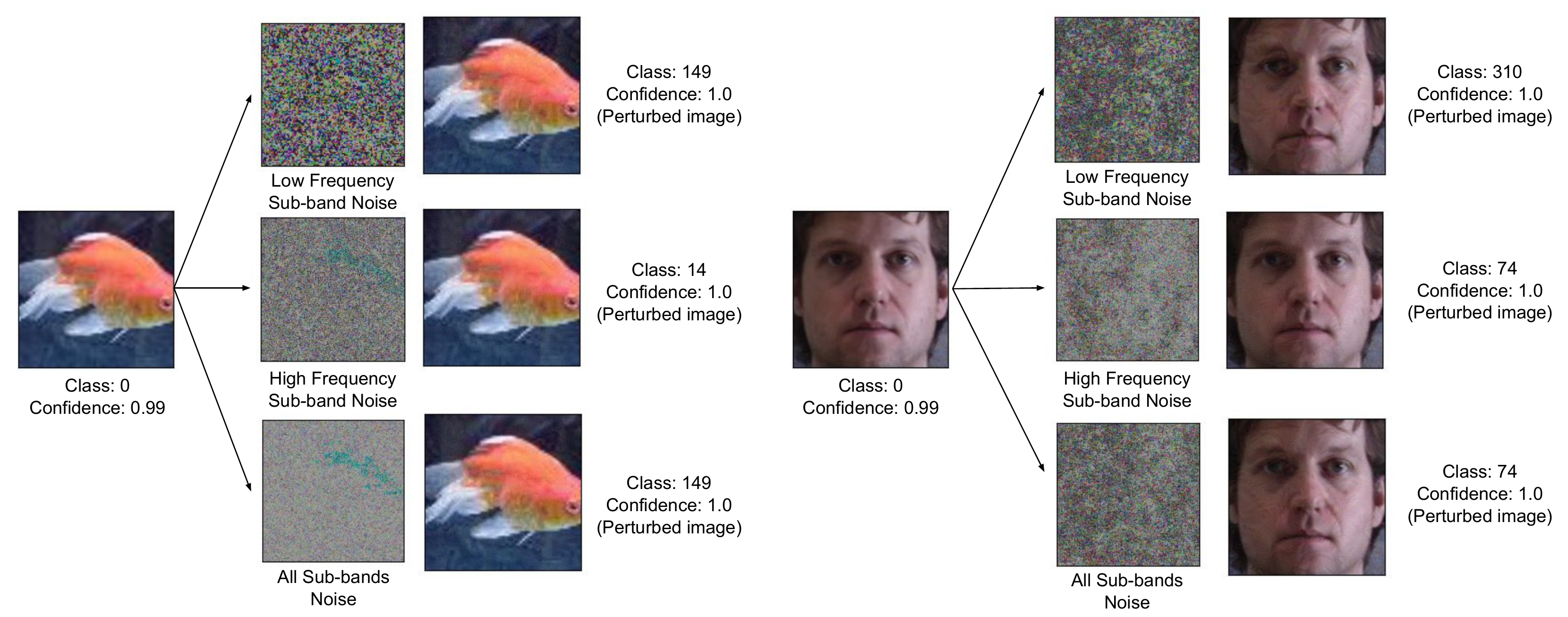}
\end{center}
\caption{Adversarial noise generated by attacking different subbands and adding to the clean image to obtain the corresponding adversarial images. Images are taken from Tiny-ImageNet \cite{yao2015tiny} (left) and Multi-PIE \cite{10.1016/j.imavis.2009.08.002} (right). It is observed that adversarial images generated using low-frequency or high-frequency or both the components, are effective in fooling the CNN with high confidence.}
\label{indiv-wavelet-Advimages}
\end{figure*}

\subsection{Effectiveness of WaveTransform}

To evaluate the effectiveness of attacking different subbands, we performed experiments with individual subbands and different combinations of subbands. Fig. \ref{fig:band-perturbed} shows samples of clean and adversarial images corresponding to individual subband from the Multi-PIE \cite{10.1016/j.imavis.2009.08.002} and Tiny-ImageNet \cite{yao2015tiny} databases. Individual wavelet components of both image classes help understand the effect of adversarial noise on each frequency information. While the noise in the low-frequency image is quasi subtle, it is visible in the high-frequency components. Among the high-frequency components, the HH component yields the highest amount of distortion. It is interesting to note that the final adversarial images are close to their clean counterpart. Fig. \ref{indiv-wavelet-Advimages} shows adversarial images generated by perturbing the different frequency components of an image. Adversarial image, whether created from low-frequency perturbation or high-frequency, can fool the classifier with high confidence.

Table \ref{tab:sb-update} summarizes the results on each database for the clean as well as the adversarial images. The ResNet-50 model trained on the CIFAR-10 database yields $94.38$\% object classification accuracy on clean test images. The performance of the model decreases drastically when any of the wavelet frequency band is perturbed. For example, when only the low frequency band is corrupted, the model fails and can classify $3.11$\% test images only. The performance drops further when all the high subbands ($HL$, $LH$, and $HH$) are perturbed and yields only $1.03$\% classification accuracy. The results show that each element is essential, and perturbing any component can significantly reduce the network performance.
Similarly, on the Tiny-ImageNet \cite{yao2015tiny}, the proposed attack can fool the ResNet-50 model almost perfectly. The model, which yields $75.29$\% object recognition accuracy on clean test images, gives $0.01$\% accuracy on adversarial images. On the Multi-PIE database, the ResNet-50 model yields $99.41$\% face identification accuracy, which reduces to $0.06$\% when both low and high-frequency components are perturbed.

\begin{table}[!t]
\caption{Classification rates (\%) of the original images and adversarial images generated by attacking different wavelet subbands. The ResNet-50 model is used for CIFAR-10 \cite{Krizhevsky2009LearningML}, Multi-PIE \cite{10.1016/j.imavis.2009.08.002} and Tiny-ImageNet \cite{yao2015tiny}. The results on F-MNIST \cite{xiao2017fashionmnist} are reported using custom CNN (refer Table \ref{tab:fmnist-model}). Bold values represent the best fooling rate achieved by perturbing all subbands, and `underline' value represents if the fooling rate is the same with all subbands perturbation.}
    \label{tab:sb-update}
    \begin{center}
            \begin{tabular}{|l|c|c|c|c|}
\hline
\hspace{2pt}\textbf{Dataset}\hspace{2pt} & \hspace{2pt}\textbf{CIFAR-10}\hspace{2pt} & \hspace{2pt}\textbf{F-MNIST}\hspace{2pt} & \hspace{2pt}\textbf{Multi-PIE}\hspace{2pt}  & \hspace{2pt}\textbf{Tiny-ImageNet}\hspace{2pt} \\ \hline\hline
\begin{tabular}[l]{@{}l@{}}\hspace{2pt}Original Accuracy\end{tabular}    & \textbf{94.38}    & \textbf{87.88} & \textbf{99.41}      & \textbf{75.29}         \\ \hline
\begin{tabular}[l]{@{}l@{}}\hspace{2pt}LL Subband Attack\end{tabular}    & 3.11     & 59.04 & 0.08       & \underline{0.01}    \\ \hline
\begin{tabular}[l]{@{}l@{}}\hspace{2pt}LH Subband Attack\end{tabular}    & 7.10     & 78.51 & \underline{0.06} & \underline{0.01}    \\ \hline
\begin{tabular}[l]{@{}l@{}}\hspace{2pt}HL Subband Attack\end{tabular}    & 6.56     & 72.73 & 0.10        & \underline{0.01}    \\ \hline
\begin{tabular}[l]{@{}l@{}}\hspace{2pt}HH Subband Attack\end{tabular}    & 13.77    & 80.56 & 0.10        & 0.54          \\ \hline
\begin{tabular}[l]{@{}l@{}}\hspace{2pt}High Subbands Attack \hspace{2pt}\end{tabular} & 1.03     & 70.04 & 0.08       & \underline{0.01}    \\ \hline
\begin{tabular}[l]{@{}l@{}}\hspace{2pt}All Subbands Attack\end{tabular} & \textbf{0.16} & \textbf{58.36} & \textbf{0.06} & \textbf{0.01} \\ \hline
\end{tabular}
        \end{center}
     
\end{table}

On the Fashion-MNIST \cite{xiao2017fashionmnist} database, the proposed attack reduces the model accuracy from $87.88$\% to $58.36$\%. In comparison to other databases, the drop on the F-MNIST database is low, which can be attributed to the lack of high textural and object shape information. It is also interesting to note that the model used on F-MNIST is much shallower as compared to the models used for other databases. While the deeper models give higher recognition accuracy as compared to the shallow model; they also find more sensitivity against adversarial perturbations in comparison to the shallow model \cite{kurakin2016adversarial+}. The results reported in Table \ref{tab:sb-update} corresponds to a \textit{`white-box'} scenario where an attacker has complete access to the classification network.

\noindent \textbf{Importance of Filter:} A filter is a critical part of DWT; therefore, to understand which types of filters are useful in crafting the proposed attack, we have performed experiments with multiple types of filters: Haar, Daubechies (db2 and db3), and Bi-orthogonal. Across the experiments on each database and CNN models, it is observed that `Haar' is more effective in comparison to other filters in reducing the classification performance. For example, on the F-MNIST \cite{xiao2017fashionmnist} database, the Haar filter reduces the model accuracy to $58.36\%$ from $87.88$\%, which is at least $1.61$\% better than Daubechies and Bi-orthogonal.


\begin{figure}[!t]
\begin{center}
   \includegraphics[width=0.75\linewidth]{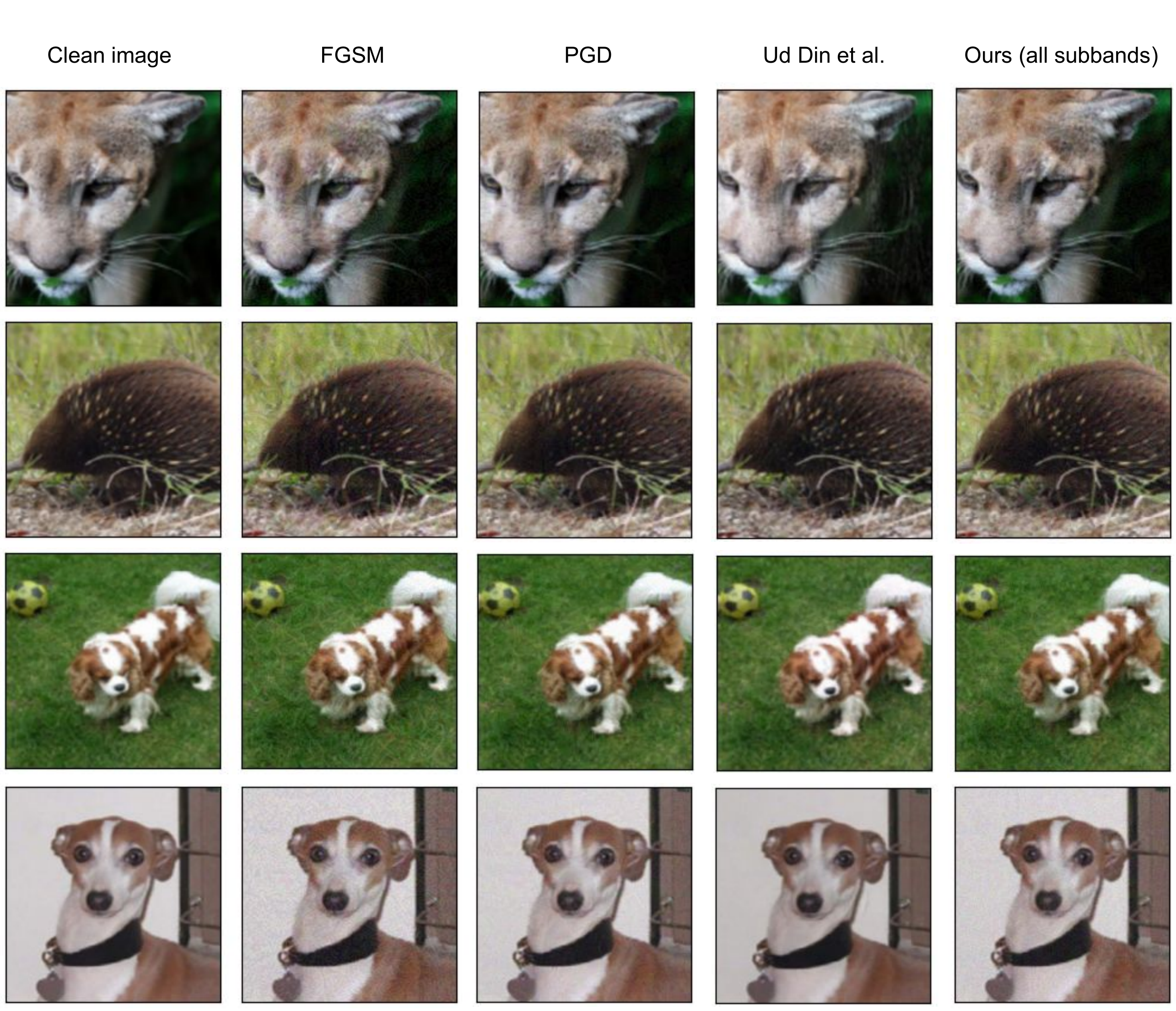}
\end{center}
\caption{Comparison of the adversarial images generated using the proposed and existing attacks including FGSM, PGD and Ud Din et al. \cite{UDDIN2020146} on the ImageNet. Images used have been resized and center-cropped to make them of size $224\times{224}\times{3}$.}
\label{fig:collage}
\end{figure}

\subsection{Comparison with Existing Attack Algorithms}

We next compare the performance of WaveTransform with pixel-level attacks and recent wavelet based attacks in literature. Fig. \ref{fig:collage} shows the adversarial images generated using the proposed, existing pixel level attacks FGSM and PGD, and steganography attack by Ud Din et al. \cite{UDDIN2020146}. 

\noindent \textbf{Pixel-level Attacks:} While most of the existing adversarial attack algorithms work at the pixel level, i.e., in the image space only; the proposed attack works at the transformation level. Therefore, we have also compared the performance of the proposed attack with popular methods such as Projected Gradient Descent (PGD) \cite{aleks2017deep} and Fast Gradient Sign Method (FGSM) \cite{goodfellow2014explaining} with $\epsilon=0.03$ in terms of accuracy and image degradation. Image degradation metrics such as Universal Image Quality Index (UIQI) \cite{995823} is a useful measure for attack quality. An adversarial example with a higher UIQI (with the maximum being 100, for the original image), is perceptually harder to distinguish from the clean image. On the CIFAR-10 database, while the proposed attack with perturbation on both low and high-frequency subbands reduces the performance of ResNet-50 to $0.16$\% from $94.38$\%, existing PGD and FGSM reduce the performance to $0.06$\% and $46.27$\%, respectively. Similarly, on the ImageNet validation database, the proposed attack reduces the performance of ResNet-50 to $0.05$\% from $76.13$\%. On the contrary, the existing PGD and FGSM attacks reduce the recognition accuracy to $0.01$\% and $8.2$\%, respectively. The experiments show that the proposed attack can either surpass the existing attack or perform comparably on both databases.

While the perturbation strength both in the existing and proposed attacks is fixed to quasi imperceptible level, we have evaluated the image quality of the adversarial examples. The average UIQI computed from the adversarial examples computed on the CIFAR-10 and ImageNet databases show a value of more than 99. The higher value (close to maximum, i.e., 100) shows that both existing and proposed attacks retain the quality of images and make the noise imperceptible to humans.

\noindent \textbf{Comparison with Recent Attack:} The closest attack to the proposed attack is recently proposed by Yahya et al. \cite{9000814} and Ud Din et al. \cite{UDDIN2020146}. These attacks are based on the concept of steganography, where a watermark image referred to as a secret image is embedded in the clean images using wavelet decomposition. The performance of the model is dependent on the secret image. To make the attack highly successful, i.e., to reduce the CNN's recognition performance, a compelling steganography image is selected based on its fooling rate on the target CNN. However, the proposed approach has no requirement of an additional watermark image and learns the noise vector from the network itself. Since Yahya et al. \cite{9000814} have shown the effectiveness of the attack on the simple MNIST database only, we have compared the performance with Ud Din et al. \cite{UDDIN2020146}. They have evaluated their method on a validation set of ImageNet. 

To maintain consistency, the experiments are performed on a validation set of ImageNet with ResNet-50. Along with visual comparison, the results are also compared using fooling ratio as the evaluation metric, which is defined as 
\begin{equation}
    \psi=\frac{|\{f( \,x_{i}+\eta) \,\neq f( \,x_{i}) \,\}|}{M} \,, \forall i \in \{1,2,...,M\}
\end{equation}
where $f$ is a trained classifier, $x_{i}$ is a clean image from the database, $M$ is the total number of samples, and $\eta$ is the adversarial noise. Using the best steganography image, the attack by Ud Din et al. \cite{UDDIN2020146} on a pretrained ResNet-50 achieves a fooling ratio of $84.77$\% whereas, the proposed attack achieves a fooling ratio of $99.95$\%.



\begin{figure}[!t]
\centering
\includegraphics[width=0.7\textwidth]{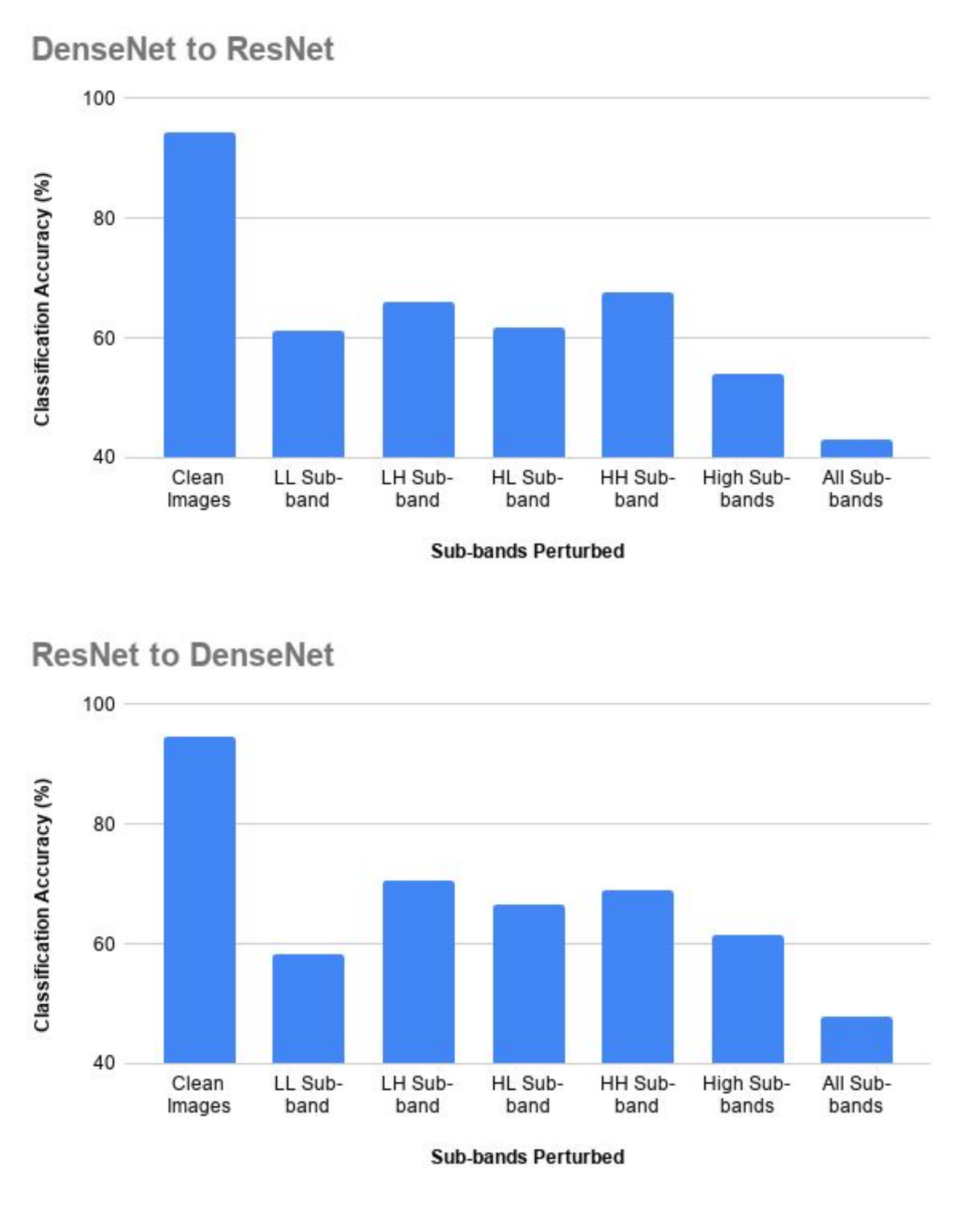}
\caption{Illustrating transfer capability of the proposed attack using CIFAR-10 database \cite{Krizhevsky2009LearningML}. The graph on the right shows the results of adversarial images generated on ResNet-50 being tested on DenseNet-121. The plot on the left shows the results of adversarial images generated on DenseNet-121 being tested on ResNet-50. The performance of ResNet-50 and DenseNet-121 are degraded upto $43.02$\% from $94.38$\% and $47.82$\% from $94.76$\%, respectively.}
\label{tranfer-results}
\end{figure}

\subsection{Transferability and Resiliency of WaveTransform}

Finally, we evaluate the transferability and resiliency of the proposed attack on multiple databases. 

\noindent \textbf{Transferability:} In the real-world settings, the attacker might not know the target CNN model, which he/she wants to fool. In such a scenario, to make the attack more practical, it is necessary to evaluate its effectiveness with an unseen testing network - the adversarial images generated using one model are used to fool another unseen model. The scenario refers to \textit{`black-box'} setting in adversarial attack literature \cite{akhtar2018threat} where an attacker does not have access to the target model. The experiments are performed on the CIFAR-10 \cite{Krizhevsky2009LearningML} and Multi-PIE \cite{10.1016/j.imavis.2009.08.002} databases. 

For CIFAR-10 \cite{Krizhevsky2009LearningML}, two state-of-the-art CNN models are used, i.e., ResNet-50 and DenseNet-121, and the results are summarized in Fig. \ref{tranfer-results}. The ResNet model yields $94.38$\% accuracy on clean images of CIFAR-10 \cite{Krizhevsky2009LearningML}; on the other hand, DenseNet gives $94.76$\% classification accuracy. When the adversarial images generated using the ResNet model are used for classification, the performance of the DenseNet model reduces to $46.94$\%. Similar performance reduction can be observed on the performance of the ResNet model when the adversarial images generated using the DenseNet model are used. The adversarial images generated by perturbing all the high-frequency wavelet bands reduce the classification accuracy up to $51.36$\%. The sensitivity of the network against the unseen attack generated models shows the practicality of the proposed attack. 
Other than that, when adversarial examples are generated using the ResNet on Multi-PIE \cite{10.1016/j.imavis.2009.08.002} and used for classification by InceptionNet \cite{szegedy2014going}, the performance of the network reduces by $26.77$\%. The perturbation of low-frequency components hurts the performance most in comparison to the modification of high-frequency components. The highest reduction in accuracy across the unseen testing network is observed when both low and high-frequency components are perturbed. 

WaveTransform works by corrupting low frequency and high frequency information contained in the image. It is well understood that the low frequency information corresponds to the high level features learned in the deeper layers of the network \cite{matthew2014visualizing,geirhos2018imagenet}. Moosavi-Dezfooli et al. \cite{moosavi2017universal} have shown that the high level features learned by different models tend to be similar. We assert that since the proposed method perturbs low frequency information that is used across models, it shows good transferability.

\begin{table}[t]
\caption{Classification rates (\%) for the original and adversarial images generated by attacking different wavelet subbands in the presence of kernel defense \cite{wang2019high}. ResNet-50 model is used for CIFAR-10 \cite{Krizhevsky2009LearningML} and the results on F-MNIST \cite{xiao2017fashionmnist} are reported using the custom CNN.} 
    \label{tab:sb-update-kernel}
\begin{center}

\begin{tabular}{|c|c|c|c|c|c|c|c|c|} \hline

\multirow{2}{*}{\textbf{Database}} & \multirow{2}{*}{} & \multirow{2}{*}{\hspace{2pt}\textbf{Original}\hspace{2pt}} & \multicolumn{6}{c|}{\textbf{Wavelet Subbands}} \\ \cline{4-9}
 &  &  & \textbf{LL} & \textbf{LH} & \textbf{HL} & \textbf{HH} & \textbf{High} & \textbf{All} \\  \hline\hline

\multirow{2}{*}{\hspace{2pt}CIFAR-10\hspace{2pt}} & \begin{tabular}[c]{@{}l@{}}\hspace{2pt}Before Defense\hspace{2pt}\end{tabular} & \textbf{94.38} & 3.11	& 7.10	& 6.56	& 13.77	& 1.03	& \textbf{0.16} \\ \cline{2-9}
 & \begin{tabular}[c]{@{}l@{}}After Defense\end{tabular} & \textbf{91.92} & 2.42 & 5.73 & 5.03 & 10.05 & 0.65 & \textbf{0.11} \\ \hline
 
 \multirow{2}{*}{F-MNIST} & \begin{tabular}[c]{@{}l@{}}Before Defense\end{tabular} & \textbf{87.88} & 59.04 & 78.51 & 72.73 & 80.56 & 70.04 & \textbf{58.36} \\ \cline{2-9}
 & \begin{tabular}[c]{@{}l@{}}After Defense\end{tabular} & \textbf{81.29} & \hspace{2pt}57.99\hspace{2pt} & \hspace{2pt}72.74\hspace{2pt} & \hspace{2pt}69.05 \hspace{2pt}& \hspace{2pt}74.84 \hspace{2pt}& \hspace{2pt}66.76\hspace{2pt} & \hspace{2pt}\textbf{57.84}\hspace{2pt} \\ \hline
 
\end{tabular}
\end{center}

\end{table}

\noindent \textbf{Adversarial Resiliency:} With the advancement in the adversarial attack domain, researchers have proposed several defense algorithms \cite{agarwal2018robustness,agarwal2020sign,goel2019robustness2,goel2019robustness,goel2020robustness,ren2020adversarial}. We next evaluate the resiliency of the attack images generated using the proposed WaveTransform against the recently proposed defense algorithm by Wang et al. \cite{wang2019high}\footnote{Original codes provided by the authors are used to perform the experiment.}. The concept of the defense algorithm is close to the proposed attack, thus making it a perfect fit for evaluation. The defense algorithm performs smoothing of the CNN neurons at earlier layers to reduce the effect of adversarial noise. 

Table \ref{tab:sb-update-kernel} summarizes the results with the defense algorithm on CIFAR10 and F-MNIST databases. Interestingly, we observe that in the presence of the defense algorithm, the performance of the network is further reduced. We hypothesize that, while the proposed attack is perturbing the frequency components, the kernel smoothing further attenuates the noise and yields a higher fooling rate. This phenomenon can also be seen from the accuracy of clean images. For example, the ResNet model without defense yields $94.38$\% accuracy on CIFAR-10, which reduces to $91.92$\% after defense incorporation. The proposed attack can fool the defense algorithm on each database. For example, on the CIFAR-10 database, the proposed attack reduces the accuracy up to $0.16$\% and it further reduces to $0.11$\% after the defense. Similar resiliency is observed on the F-MNIST database as well. 

The PGD attack, which shows a similar reduction in the performance on the CIFAR-10 database, is found less resilient against the defense proposed by Wang et al. \cite{wang2019high}. The defense algorithm can successfully boost the recognition performance of ResNet-50 by $30$\%; whereas, the proposed attack is found to be resilient against the defense. 

The robustness of the proposed attack is also evaluated against state-of-the-art defense methods such as Madry et al. \cite{aleks2017deep} and Zhang et al. \cite{zhang2019theoretically} on the CIFAR-10 database. The defense model presented by Madry et al. \cite{aleks2017deep} utilizes the ResNet-50 model, which yields $87.03$\% accuracy on clean images of the database, but the accuracy significantly reduces to $60.81$\% when the proposed attack is applied. Similarly, the accuracy of the defended WideResNet \cite{zagoruyko2016wide} model by Zhang et al. \cite{zhang2019theoretically} reduces to $62.73$\% from $84.92$\%.

\section{Conclusion}

High and low frequency components present in an image play a vital role when they are processed by deep learning models. Several recent research works have also highlighted that CNN models are highly sensitive towards high and low-frequency components. The attack generation algorithms in the literature generally learn the additive noise vector without considering the individual frequency components. In this research, intending to understand the role of different frequencies, we have proposed a novel attack by decomposing the images using discrete wavelet transform and adding learned adversarial noise in different frequency subbands. The experiments using multiple databases and deep learning models show that the proposed attack poses a significant challenge to the classification models. The proposed attack is further evaluated under unseen network training-testing settings to showcase its real-world application. Other than that, the proposed WaveTransform attack is found to be challenging to mitigate/defend. 

\section*{Acknowledgements}
A. Agarwal was partly supported by the Visvesvaraya PhD Fellowship. R. Singh and M. Vatsa are partially supported through a research grant from MHA, India. M. Vatsa is also partially supported through Swarnajayanti Fellowship by the  Government of India.



%
%
\bibliographystyle{splncs04}
\bibliography{27}

\begin{thebibliography}{10}
\providecommand{\url}[1]{\texttt{#1}}
\providecommand{\urlprefix}{URL }
\providecommand{\doi}[1]{https://doi.org/#1}

\bibitem{agarwal2018robustness}
Agarwal, A., Singh, R., Vatsa, M., Ratha, N.: Are image-agnostic universal
  adversarial perturbations for face recognition difficult to detect? IEEE BTAS
  pp.~1--7 (2018)

\bibitem{agarwal2020sign}
Agarwal, A., Vatsa, M., Singh, R.: The role of sign and direction of gradient
  on the performance of {CNN}. IEEE CVPRW  (2020)

\bibitem{agarwal2020noise}
Agarwal, A., Vatsa, M., Singh, R., Ratha, N.: Noise is inside me! generating
  adversarial perturbations with noise derived from natural filters. IEEE CVPRW
   (2020)

\bibitem{akhtar2018threat}
Akhtar, N., Mian, A.: Threat of adversarial attacks on deep learning in
  computer vision: A survey. IEEE Access  \textbf{6},  14410--14430 (2018)

\bibitem{athalye2017synthesizing}
Athalye, A., Engstrom, L., Ilyas, A., Kwok, K.: Synthesizing robust adversarial
  examples. ICML pp. 284--293 (2018)

\bibitem{behzadan2017vulnerability}
Behzadan, V., Munir, A.: Vulnerability of deep reinforcement learning to policy
  induction attacks. In: MLDM. pp. 262--275. Springer (2017)

\bibitem{Cao18}
Cao, Q., Shen, L., Xie, W., Parkhi, O.M., Zisserman, A.: Vggface2: A dataset
  for recognising faces across pose and age. In: IEEE FG. pp. 67--74 (2018)

\bibitem{carlini2017towards}
Carlini, N., Wagner, D.: Towards evaluating the robustness of neural networks.
  In: IEEE S\&P. pp. 39--57 (2017)

\bibitem{carlini2018audio}
Carlini, N., Wagner, D.: Audio adversarial examples: Targeted attacks on
  speech-to-text. In: IEEE S\&PW. pp.~1--7 (2018)

\bibitem{chen2018ead}
Chen, P.Y., Sharma, Y., Zhang, H., Yi, J., Hsieh, C.J.: {EAD}: elastic-net
  attacks to deep neural networks via adversarial examples. In: AAAI (2018)

\bibitem{deng2009imagenet}
{Deng}, J., {Dong}, W., {Socher}, R., {Li}, L., and: Imagenet: A large-scale
  hierarchical image database. IEEE CVPR pp. 710--719 (2009)

\bibitem{UDDIN2020146}
Din, S.U., Akhtar, N., Younis, S., Shafait, F., Mansoor, A., Shafique, M.:
  Steganographic universal adversarial perturbations. Pattern Recognition
  Letters  \textbf{135},  146 -- 152 (2020)

\bibitem{dong2018boosting}
Dong, Y., Liao, F., Pang, T., Su, H., Zhu, J., Hu, X., Li, J.: Boosting
  adversarial attacks with momentum. In: IEEE CVPR. pp. 9185--9193 (2018)

\bibitem{gao2018black}
Gao, J., Lanchantin, J., Soffa, M.L., Qi, Y.: Black-box generation of
  adversarial text sequences to evade deep learning classifiers. In: IEEE
  S\&PW. pp. 50--56 (2018)

\bibitem{geirhos2018imagenet}
Geirhos, R., Rubisch, P., Michaelis, C., Bethge, M., Wichmann, F.A., Brendel,
  W.: Imagenet-trained cnns are biased towards texture; increasing shape bias
  improves accuracy and robustness. ICLR  (2019)

\bibitem{goel2019robustness2}
Goel, A., Agarwal, A., Vatsa, M., Singh, R., Ratha, N.: Deep{R}ing: Protecting
  deep neural network with blockchain. IEEE CVPRW  (2019)

\bibitem{goel2019robustness}
Goel, A., Agarwal, A., Vatsa, M., Singh, R., Ratha, N.: Securing {CNN} model
  and biometric template using blockchain. IEEE BTAS pp.~1--6 (2019)

\bibitem{goel2020robustness}
Goel, A., Agarwal, A., Vatsa, M., Singh, R., Ratha, N.: {DNDN}et: Reconfiguring
  {CNN} for adversarial robustness. IEEE CVPRW  (2020)

\bibitem{goel2018smartbox}
Goel, A., Singh, A., Agarwal, A., Vatsa, M., Singh, R.: Smartbox: Benchmarking
  adversarial detection and mitigation algorithms for face recognition. IEEE
  BTAS  (2018)

\bibitem{goodfellow2014explaining}
Goodfellow, I.J., Shlens, J., Szegedy, C.: Explaining and harnessing
  adversarial examples. arXiv preprint arXiv:1412.6572  (2014)

\bibitem{aaa2018goswami}
Goswami, G., Ratha, N., Agarwal, A., Singh, R., Vatsa, M.: Unravelling
  robustness of deep learning based face recognition against adversarial
  attacks. AAAI pp. 6829--6836 (2018)

\bibitem{ijcv2019goswami}
Goswami, G., Agarwal, A., Ratha, N., Singh, R., Vatsa, M.: Detecting and
  mitigating adversarial perturbations for robust face recognition.
  International Journal of Computer Vision  \textbf{127},  719--742 (2019),
  doi: \url{10.1007/s11263-019-01160-w}

\bibitem{10.1016/j.imavis.2009.08.002}
Gross, R., Matthews, I., Cohn, J., Kanade, T., Baker, S.: Multi-pie. I\&V Comp.
   \textbf{28}(5),  807--813 (2010)

\bibitem{he2015deep}
He, K., Zhang, X., Ren, S., Sun, J.: Deep residual learning for image
  recognition. In: IEEE CVPR. pp. 770--778 (2016)

\bibitem{huang2016densely}
Huang, G., Liu, Z., Van Der~Maaten, L., Weinberger, K.Q.: Densely connected
  convolutional networks. In: IEEE CVPR. pp. 4700--4708 (2017)

\bibitem{kiefer1952}
Kiefer, J., Wolfowitz, J., et~al.: Stochastic estimation of the maximum of a
  regression function. The Annals of Mathematical Statistics  \textbf{23}(3),
  462--466 (1952)

\bibitem{kingma2014adam}
Kingma, D.P., Ba, J.: Adam: A method for stochastic optimization. ICLR  (2015)

\bibitem{Krizhevsky2009LearningML}
Krizhevsky, A.: Learning multiple layers of features from tiny images (2009)

\bibitem{kurakin2016adversarial}
Kurakin, A., Goodfellow, I., Bengio, S.: Adversarial examples in the physical
  world. ICLR-W  (2017)

\bibitem{kurakin2016adversarial+}
Kurakin, A., Goodfellow, I., Bengio, S.: Adversarial machine learning at scale.
  ICLR  (2017)

\bibitem{aleks2017deep}
Madry, A., Makelov, A., Schmidt, L., Tsipras, D., Vladu, A.: Towards deep
  learning models resistant to adversarial attacks. ICLR  (2018)

\bibitem{matthew2014visualizing}
Matthew, D., Fergus, R.: Visualizing and understanding convolutional neural
  networks. In: ECCV. pp. 6--12 (2014)

\bibitem{moosavi2017universal}
Moosavi-Dezfooli, S.M., Fawzi, A., Fawzi, O., Frossard, P.: Universal
  adversarial perturbations. In: IEEE CVPR. pp. 1765--1773 (2017)

\bibitem{moosavi2016deepfool}
Moosavi-Dezfooli, S.M., Fawzi, A., Frossard, P.: Deepfool: a simple and
  accurate method to fool deep neural networks. In: IEEE CVPR. pp. 2574--2582
  (2016)

\bibitem{ren2020adversarial}
Ren, K., Zheng, T., Qin, Z., Liu, X.: Adversarial attacks and defenses in deep
  learning. Elsevier Engineering pp. 1--15 (2020)

\bibitem{ILSVRC15}
Russakovsky, O., Deng, J., Su, H., Krause, J., Satheesh, S., Ma, S., Huang, Z.,
  Karpathy, A., Khosla, A., Bernstein, M., Berg, A.C., Fei-Fei, L.: {ImageNet
  Large Scale Visual Recognition Challenge}. IJCV  \textbf{115}(3),  211--252
  (2015)

\bibitem{singh2020robustness}
Singh, R., Agarwal, A., Singh, M., Nagpal, S., Vatsa, M.: On the robustness of
  face recognition algorithms against attacks and bias. AAAI SMT  (2020)

\bibitem{szegedy2014going}
Szegedy, C., Liu, W., Jia, Y., Sermanet, P., Reed, S., Anguelov, D., Erhan, D.,
  Vanhoucke, V., Rabinovich, A.: Going deeper with convolutions. In: IEEE CVPR.
  pp.~1--9 (2015)

\bibitem{wang2019high}
Wang, H., Wu, X., Huang, Z., Xing, E.P.: High-frequency component helps explain
  the generalization of convolutional neural networks. In: IEEE/CVF CVPR. pp.
  8684--8694 (2020)

\bibitem{xiang2019generating}
Xiang, C., Qi, C.R., Li, B.: Generating 3d adversarial point clouds. In: IEEE
  CVPR. pp. 9136--9144 (2019)

\bibitem{xiao2018generating}
Xiao, C., Li, B., Zhu, J.Y., He, W., Liu, M., Song, D.: Generating adversarial
  examples with adversarial networks. arXiv preprint arXiv:1801.02610  (2018)

\bibitem{xiao2017fashionmnist}
Xiao, H., Rasul, K., Vollgraf, R.: Fashion-mnist: a novel image dataset for
  benchmarking machine learning algorithms. arXiv preprint arXiv:1708.07747
  (2017)

\bibitem{xie2017adversarial}
Xie, C., Wang, J., Zhang, Z., Zhou, Y., Xie, L., Yuille, A.: Adversarial
  examples for semantic segmentation and object detection. In: IEEE ICCV. pp.
  1369--1378 (2017)

\bibitem{9000814}
{Yahya}, Z., {Hassan}, M., {Younis}, S., {Shafique}, M.: Probabilistic analysis
  of targeted attacks using transform-domain adversarial examples. IEEE Access
  \textbf{8},  33855--33869 (2020)

\bibitem{yao2015tiny}
Yao, L., Miller, J.: Tiny imagenet classification with convolutional neural
  networks. CS 231N  \textbf{2}(5), ~8 (2015)

\bibitem{yi2014learning}
Yi, D., Lei, Z., Liao, S., Li, S.Z.: Learning face representation from scratch.
  arXiv preprint arXiv:1411.7923  (2014)

\bibitem{8611298}
{Yuan}, X., {He}, P., {Zhu}, Q., {Li}, X.: Adversarial examples: Attacks and
  defenses for deep learning. IEEE TNNLS  \textbf{30}(9),  2805--2824 (2019)

\bibitem{zagoruyko2016wide}
Zagoruyko, S., Komodakis, N.: Wide residual networks. arXiv preprint
  arXiv:1605.07146  (2016)

\bibitem{zhang2019theoretically}
Zhang, H., Yu, Y., Jiao, J., Xing, E.P., Ghaoui, L.E., Jordan, M.I.:
  Theoretically principled trade-off between robustness and accuracy. ICML
  (2019)

\bibitem{995823}
{Zhou Wang}, {Bovik}, A.C.: A universal image quality index. IEEE Signal
  Processing Letters  \textbf{9}(3),  81--84 (2002)

\end{thebibliography}
\end{document}